\documentclass[11pt]{article}
\usepackage{geometry}
\usepackage{coling2020}
\usepackage{times}
\usepackage{graphicx}
\usepackage{url}
\usepackage{latexsym}
\usepackage{multirow}
\usepackage{microtype}
\usepackage{epstopdf}
\usepackage{caption}

\usepackage{subfigure}

\colingfinalcopy 

\title{YNU-HPCC at SemEval-2020 Task 8: Using a Parallel-Channel Model for Memotion Analysis  }

\author{Li Yuan, Jin Wang and Xuejie Zhang \\
  School of Information Science and Engineering \\
  Yunnan University \\
  Kunming, China \\
  {\tt Contact:\{wangjin, xjzhang\}@ynu.edu.cn} \\}

\date{}

\begin{document}
\maketitle
\begin{abstract}
 In recent years, the growing ubiquity of Internet memes on social media platforms, such as Facebook, Instagram, and Twitter, has become a topic of immense interest. However, the classification and recognition of memes is much more complicated than that of social text since it involves visual cues and language understanding. To address this issue, this paper proposed a parallel-channel model to process the textual and visual information in memes and then analyze the sentiment polarity of memes. In the shared task of identifying and categorizing memes, we preprocess the dataset according to the language behaviors on social media. Then, we adapt and fine-tune the Bidirectional Encoder Representations from Transformers (BERT), and two types of convolutional neural network models (CNNs) were used to extract the features from the pictures. We applied an ensemble model that combined the BiLSTM, BIGRU, and Attention models to perform cross domain suggestion mining. The officially released results show that our system performs better than the baseline algorithm. Our team won nineteenth place in subtask A (Sentiment Classification). The code of this paper is availabled at : \url{https://github.com/YuanLi95/Semveal2020-Task8-emotion-analysis}.
\end{abstract}
\blfootnote{
    %
    %
    \hspace{-0.65cm}  
    %
    This work is licensed under a Creative Commons Attribution 4.0 International License. License details: http://creativecommons.org/licenses/by/4.0/.
    %
    %
    %
}
\section{Introduction}

In recent years, memes that combine pictures and text have been widely used in social media. Using memes can help users to express richer meaning and emotion compared with using text or images alone; hence, it is worthwhile to analyze the sentiment expressions of memes. An example of a meme is shown in Figure 1. Moreover, recognizing and analyzing the meaning and sentiment of memes is much more difficult than analyzing social texts or pictures.
\begin{figure}[ht]
\centering
\includegraphics[scale=0.7]{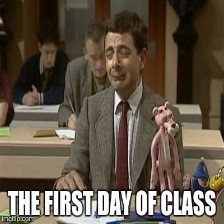}
\caption{\label{font-table}The text in this meme is “the first day of class”.}

\end{figure}

In SemEval-2020 Task 8:Memotion Analysis~\cite{chhavi2020memotion}, the organizers hoped that the task would increase the research attention given to the topic. The task is divided into three subtasks.

$\bullet$ Task A- Sentiment Classification: Given an Internet meme, the first task is to classify its sentiment polarity. The meme shown in Figure 1 is classified as very positive.

$\bullet$ Task B- Humor Classification: Given an Internet meme, the system has to identify the type of humor expressed. The meme shown in Figure 1 is classified as sarcastic, humorous, not offensive and not motivational.

$\bullet$ Task C- Scales of Semantic Classes: The third task is to quantify the extent to which a particular effect is being expressed.

Memes and this issue have attracted the attention of researchers.
In a previous study, Borth~\shortcite{Borth2013} pioneered the sentiment analysis of visual content with SentiBank. Another study implemented Optical Character Recognition (OCR) to extract the text captions of memes and then classified the sentiment polarity of the text using the Naive Bayes algorithm~\cite{Amalia2018}. For a similar meme sentiment analysis task, Zhao~\shortcite{Zhao2019} developed a multimodal sentiment analysis method for image-text posts, and their experiments showed that this method achieves excellent performance on the Flickr benchmark dataset. Hu and Flaxman~\shortcite{Hu2018} used GloVe to map the text to a high dimensional space and fine-tuned the pictures through Inception (a pretrained deep convolutional neural network).

In this paper, we propose a parallel channel model that includes a text channel, which is implemented to process the text in memes, and an image channel for image analysis. The text channel implements the BiLSTM, BiGRU, and BiLSTM with attention models. For the image channel, a multilayer CNN model and ResNet152 ~\cite{He2016} were applied to capture the image features. Then, the information in the two modalities is combined by a dense layer after concatenation. The experimental results show that our approach achieved good performance.

The remainder of this paper is organized as follows. Section 2 describes the proposed parallel channel model, and Section 3 presents the implementation details and experimental results. The conclusions of this study are presented in Section 4.

\section{Parallel Channel Model}
\label{intro}

\begin{figure}[t!]
\centering
\includegraphics[height = 0.43\textwidth,width = 0.62\textwidth]{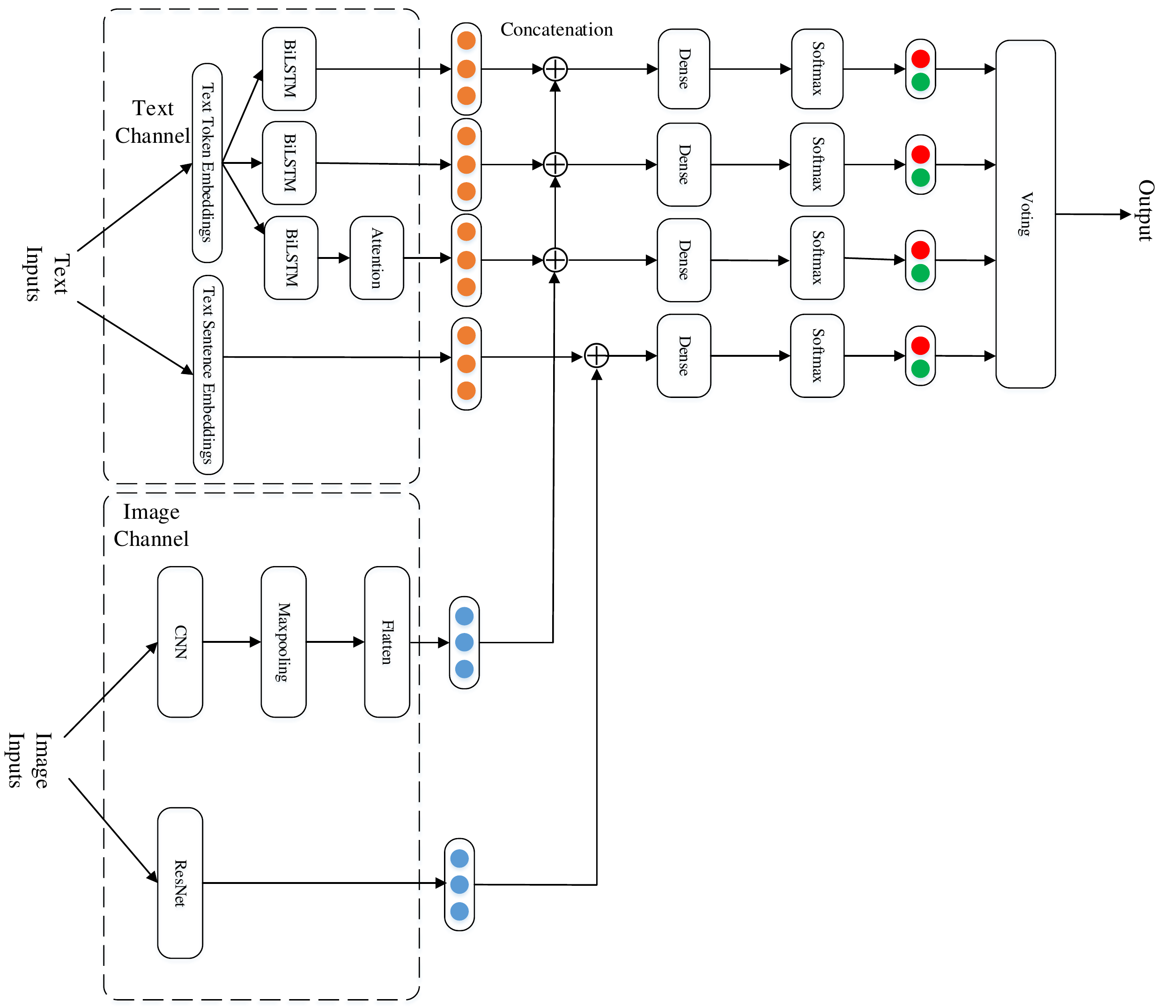}

\caption{\label{font-table}The Multimodal architecture of the parallel channel model.}
\end{figure}
\subsection{Overview}
As illustrated in Figure 2, the proposed model consists of two channels: the image channel and the text channel. We propose two different types of pretraining vectors and three different models in the text channel and two different models in the image channel as a way to extract picture features. We combined multiple models, use the soft voting mechanism, and output the results. For an input meme, ${w_s}$  represents the extracted text and ${I}$ is the image. Then, the proposed model can be expressed as follows:
\begin{equation}
\begin{array}{l}
h_i^T = f_i^T({w_s})\\
h_j^I = f_j^I(I)\\
f({w_s},I) = voting\left[ {h_i^T \oplus h_j^I} \right]
\end{array}
\end{equation}
where $i\in (1,2,3,4)$ and $j\in (1,2)$ , $f^T$ and $f^I$ represent the way to obtain special text and image features. $h_i^T$ and $h_j^I$ are the text vector and the image vector, respectively,and $f({w_s},I)$ is the final result.

\subsection{Text Channel}

{\bf Embedding Layer}. The embedding layer is the first layer of the text channel. We constructed the word vectors from a 768-dimensional BERT vector. Then, a word vector matrix was loaded into the embedding layer and then fed into different hidden layers. For longer posts, we only keep the first 128 words, which is a reasonable choice since 90\% of the posts in the dataset contain less than 128 words. We also use the sentence-level vectors from a 768-dimensional BERT vector as the text features and fed them into the fully connected layer.

\vspace{0.1cm}

\noindent {\bf Bidirectional Long Short-Term Memory (BiLSTM)}~\cite{Greff2017} is a special Recurrent Neural Network. The LSTM model can better capture the long-distance dependencies. There are various novel models based on LSTM, for instance: Wang~\shortcite{Wang2020} proposed a tree-structured regional CNN-LSTM model for valence-arousal (VA) prediction. A capsule tree LSTM model introduces a dynamic routing algorithm to construct sentence representations~\cite{Wang2019}, and experiments prove that the method improves the performance of the tree LSTM and the basic LSTM model. BiLSTM is based on LSTM and can better capture forward and backward semantic dependencies.
We show how a memory block calculates the hidden state $h_t^T$ and output ${C_t}$  using the following equations.

\begin{itemize}
\setlength{\parskip}{0pt}
\item {\bf Gate}
\begin{equation}
\begin{array}{r}
{f_t} = \sigma ({W_f}\cdot[h_{t - 1}^T,{x_t}] + {b_f})\\
{i_t} = \sigma ({W_i}\cdot[h_{t - 1}^T,{x_t}] + {b_i})\\
{o_t} = \sigma ({W_o}\cdot[h_{t - 1}^T,{x_t}] + {b_o})
\end{array}
\end{equation}

\item{\bf Transformation}
\begin{equation}
{\tilde C_t} = \tanh ({W_c}\cdot[h_{t - 1}^T,{x_t}] + {b_c})
\end{equation}

\item{\bf Status update}
\begin{equation}
\begin{array}{l}
{C_t} = {f_t} * {C_{t - 1}} + {i_t} * {{\tilde C}_t}\\
h_t^T = {o_t} * \tanh ({C_t})
\end{array}
\end{equation}
\end{itemize}
here, ${x_t}$ is the input vector; ${C_t}$ is the cell state vector; $W$ and $b$ are cell parameters; ${f_t}$ , ${i_t}$ and ${o_t}$ are gate vectors; and $\sigma $ denotes the sigmoid function.

\vspace{0.1cm}

\noindent {\bf Gated Recurrent Unit (GRU)} ~\cite{Cho2014} is a variant of LSTM that combines the forget gate and the input gate into a single update gate. It also mixes cell states and hidden states. The final model is simpler than the standard LSTM model. The effect is similar to LSTM but with fewer parameters, and it is not easy to overfit.

\vspace{0.1cm}

\noindent {\bf Attention mechanism}~\cite{Bahdanau2015} breaks the limitation that the traditional encoder-decoder structure depends on a fixed-length vector when encoding and decoding. Its implementation retains the intermediate output results of the input sequence via the LSTM encoder, trains a model to selectively learn these inputs and associates the output sequence with it when the model is output. Attention mechanisms have been widely used in various NLP fields such as the Transformer~\cite{Vaswani2017}, Neural Machine Translation~\cite{Yang2016} and aspect-level sentiment analysis~\cite{Tang2019}.

\subsection{Image Channel}

{\bf Convolutional neural networks (CNNs)}~\cite{Krizhevsky} are often used to extract image representations. A CNN is usually divided into convolution layers and pooling layers. The convolution layers are used to extract n-gram features from the picture pixels. Pooling selects a part of the input matrix and chooses the best representative for the region. The max pooling layer selects the max feature.

\vspace{0.1cm}

\noindent {\bf ResNet model}~\cite{He2016} is one of the widely used image recognition models, and it solves the deep vanishing gradient problem. The basic structure of the residual is shown in Figure 3. We used PyTorch's pretrained ResNet152 model for the feature extraction from pictures.

\begin{figure}[t!]
\centering
\includegraphics[height = 30mm,width = 0.35\textwidth,]{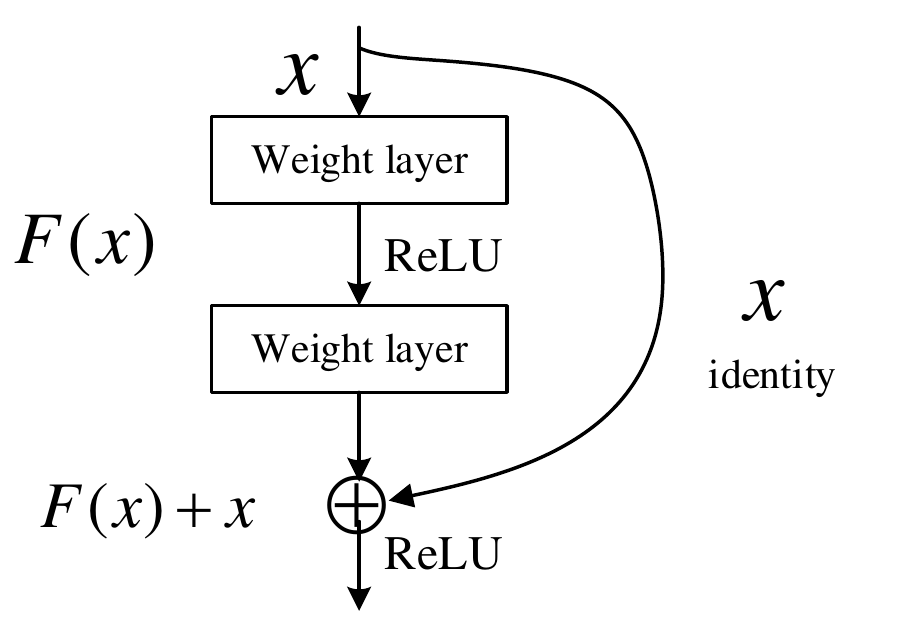}

\caption{\label{font-table} Residual learning: a building block.}
\end{figure}

\section{Experiments and Evaluation}

In this section, experiments were conducted to evaluate the proposed models on both subtasks. We also report the results of the official review. The details of the experiment are described as follows.

\subsection{Data Preparation}
\label{sect:Data}
The organizers provided 7K human annotated Internet memes labeled with semantic dimensions, namely, sentiment and the type of humor that is sarcastic, humorous, or offensive. For subtask A and subtask B, the data distributions are a little unbalanced, which make the tasks much harder. We randomly used 20\% of the memes from the provided data as the dev set to fine-tune the parameters. The Stanford tokenizer toolkit was employed to process the memes-text into an array of tokens. Meanwhile, before feeding the token array to any neural networks, they are preprocessed by following procedures:

$\bullet$ Punctuation marks, websites URLs and mailing addresses are removed,

$\bullet$ Common nonstandard expressions are restored, and

$\bullet$ Non-English letters are treated as unknown words represented by $<$unk$>$.

\subsection{Implementation Details}
\label{ssec:Details}
This experiment used Keras with the TensorFlow backend. For subtask A, we used two different pretrained word vectors, and we introduced other models. For subtask A, we tried different batch sizes and attempts, and the results are shown in Figure 4.The best batch size is 60, the best number of training epochs is 14 and the learning rate is set as 1e-5. We use Scikit-Learn to execute the grid search
~\cite{Pedregosa2011} to adjust the hyperparameters, through which we can find the best parameters for evaluating the system. The parameters given are as follows: the time step of the RNN for hidden layers 1, 2 (${h_{1,2}}$) and 3 (${h_3}$); the dimension of the dense layer ($d$); and the dropout rate ($r$). For the image channel, we also have the number of convolution layers ($c$), the number of filters ($m$), the length of the filter ($l$) and the pool ($p$). Table 1 summarizes these fine-tuned parameters.

\begin{figure}[t!]
\centering    
 
\subfigure[Fine tune of epochs] 
{
    \begin{minipage}{7cm}
    \centering          
    \includegraphics[scale=0.37]{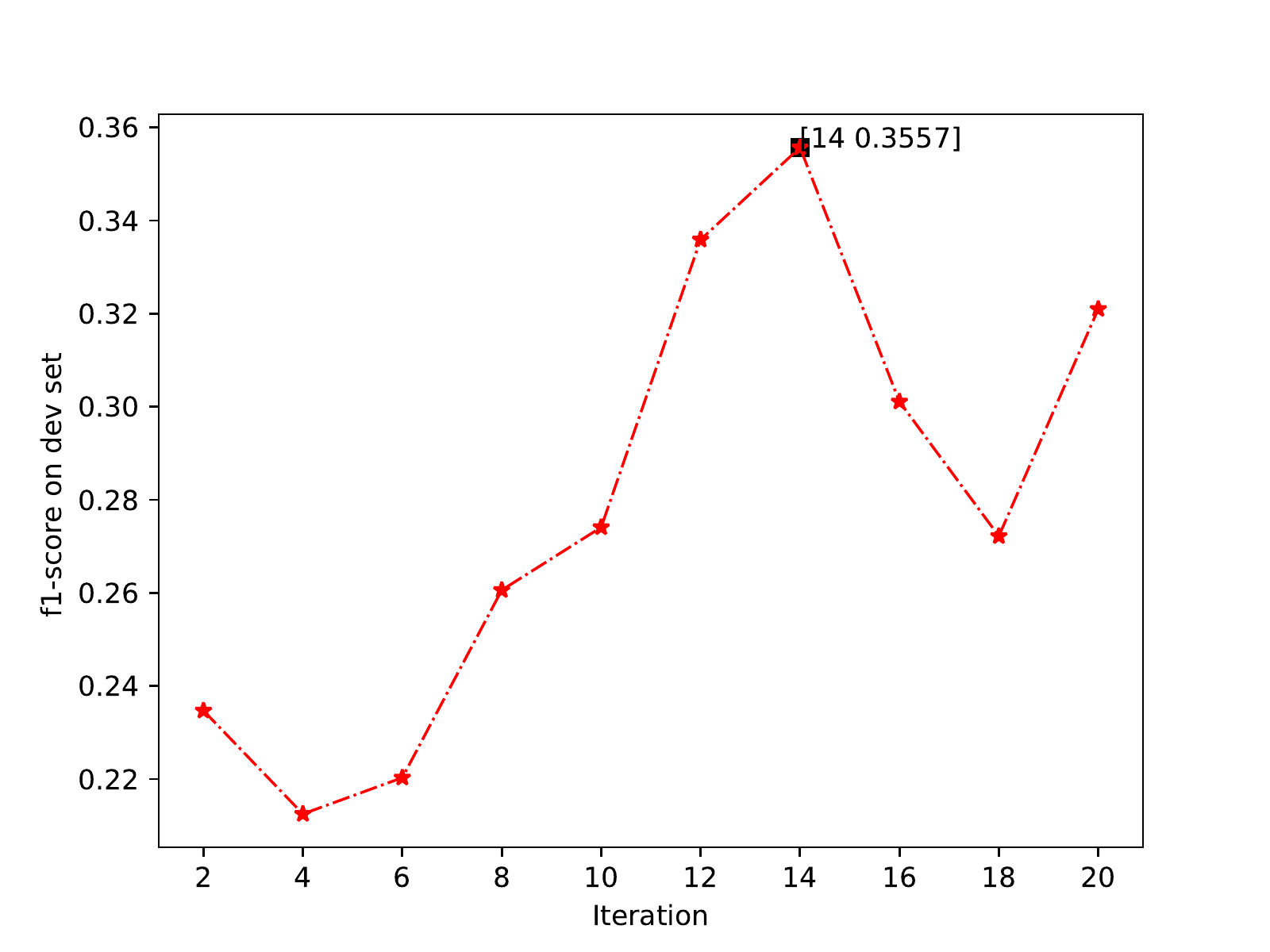}   
    \end{minipage}
}  
\subfigure[Fine tune of batch size] 
{
    \begin{minipage}{7cm}
    \centering      
    \includegraphics[scale=0.37]{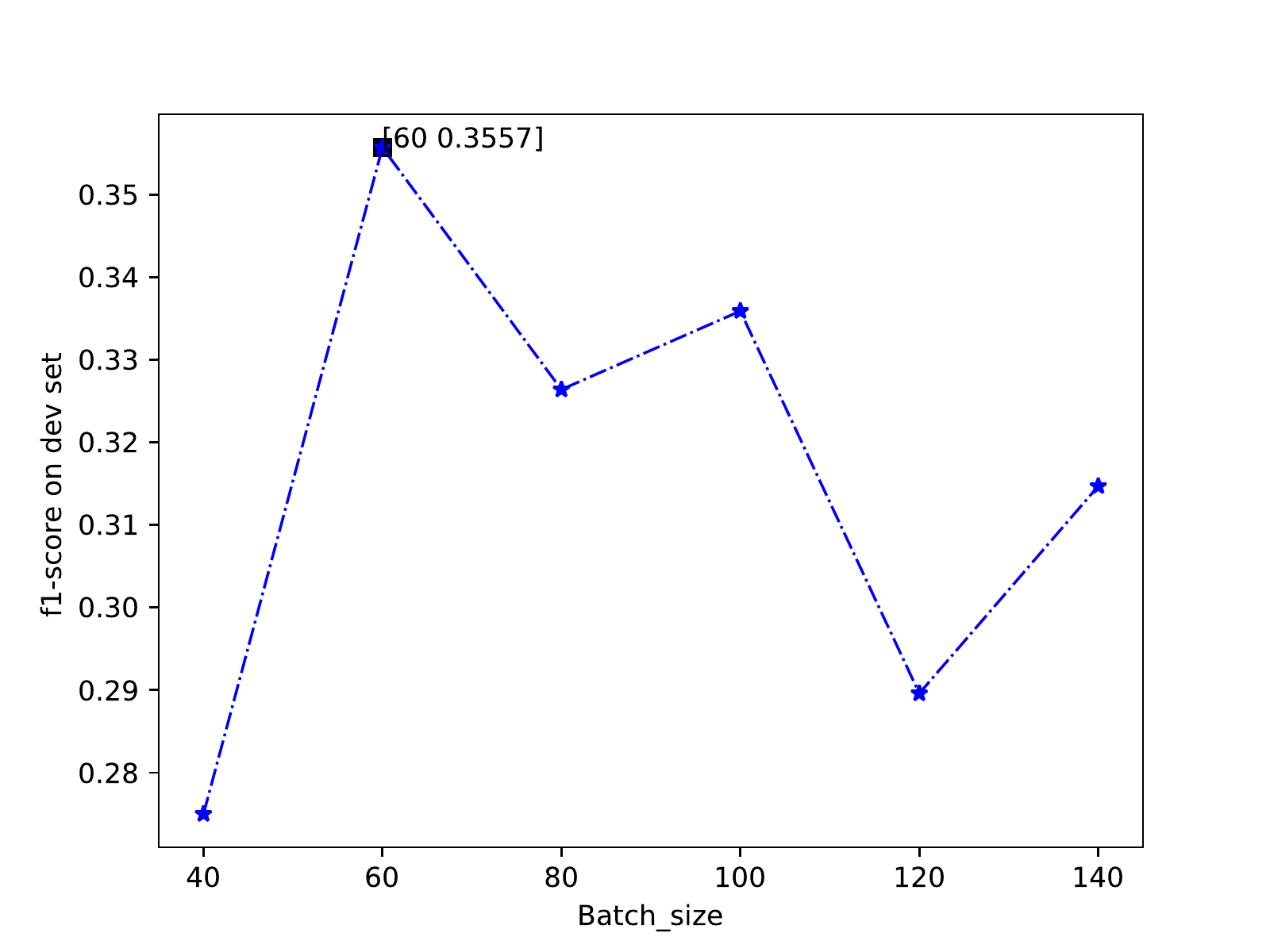}   
    \end{minipage}
}
 
\caption{\label{font-table}Fine tune of epochs and batch size.}
\label{fig:4}  
\end{figure}

\begin{table}[t!]
\begin{center}
\begin{tabular}{|c|c|c|c|c|c|}
 \hline
 \multirow{4}*{\bf Text-channel} &{\bf Model-Name}& ${h_{1,2}}$ & ${h_3}$ &$r$ &$d$\\
 \cline{2-6}
                    & BILSTM           & 300 & 160 & 0.4 & 128 \\
                    & BIGRU            & 300 & 300 & 0.3 & 128 \\
                   & BILSTM-Attention & 300 & 160 & 0.3 & 64  \\
\hline
 \multirow{2}*{\bf Picture-channel} &{\bf Model-Name}  &$c$   &$m$     &$l$   &$p$  \\\cline{2-6}
                                    & Multilayer CNN   & 6     & 64     & 3   & 0.2\\ \hline

\end{tabular}
\end{center}
\caption{\label{font-table} The best-tuned parameters for Task A. }
\end{table}

\subsection{Evaluation Metrics}

For the submission to subtask A and subtask B, its performance will evaluated based on the macro-F1 score. The F1-score is often used as an evaluation indicator of unbalanced data, and is defined as follows:

\begin{equation}
{F1 = 2 * \frac{{P*R}}{{(P + R)}}}
\end{equation}

\noindent where $P$ denotes the precision and $R$ denotes the recall. A higher F1-score indicates better classification performance.

\subsection{Results and Discussion}
\label{ssec:Results}
Table 2 shows the detailed results of the proposed our model compared to the other baseline models in ours dev set.

\begin{table}[t!]
\begin{center}
\begin{tabular}{|c|c|c|}
\hline
\multirow{2}*{\bf Model}  &\multicolumn{2}{|c|}{\bf Metrics}\\ \cline{2-3}
                                & {\bf Task A}  & {\bf Task B} \\ \hline
${\rm{BiLST}}{{\rm{M}}_{elmo}}$           &  0.286  & 0.498    \\ \hline
${\rm{BiGR}}{{\rm{U}}_{elmo}}$            &  0.304  & 0.512     \\ \hline
${\rm{BiLSTM+Attentio}}{{\rm{n}}_{elmo}}$ & 0.311   & 0.506     \\ \hline
${\rm{BiLST}}{{\rm{M}}_{bert}}$           & 0.323   & 0.523      \\ \hline
${\rm{BiGR}}{{\rm{U}}_{bert}}$            & {\bf 0.338}  & 0.531  \\ \hline
${\rm{BiLSTM+Attentio}}{{\rm{n}}_{bert}}$ & 0.328   & 0.528       \\ \hline
${\rm{Bert+ResNet}}$                       & 0.334  &  {\bf 0.536} \\ \hline
${\rm{Our}}{\rm{Mode}}{{\rm{l}}_{{\rm{ensemble}}}}$  &{\bf 0.3557}  &{\bf 0.541} \\
\hline
\end{tabular}
\end{center}
\caption{\label{font-table} The dev data experiment results.}
\end{table}

\vspace{0.1cm}

\noindent {\bf Subtask A}. Our system achieved a score that was 0.115 higher than the baseline score (0.2176). The results show that our proposed system significantly outperforms the baseline models. The main reason is that we have combined a variety of information from memes and used the BERT word embedding.

\vspace{0.1cm}

\noindent {\bf Subtask B}. Our model score was lower than the baseline score of 0.5118. We guess that it may be caused by the inconsistent data distribution between the dev set and test set, and so we need to do more research on class imbalance in the future.

\section{Conclusion}
In this paper, we describe a task system that we submitted to SemEval-2020 for Memotion Analysis. We propose a two parallel channel model. In the text channel, we use 3 RNN models and 2 types of pretraining vectors. In the image channel, we used a pretrained model and a CNN model. We participated in subtasks A and B, and obtained nineteenth place in subtask A. In future work, we will test more novel fusion methods so that the picture features can be better combined with token embedding.

\section*{Acknowledgements}
This work was supported by the National Natural Science Foundation of China (NSFC) under Grant No. 61966038, 61702443 and 61762091. The authors would like to thank the anonymous reviewers for their constructive comments.


\bibliographystyle{coling}
\bibliography{semeval2020}

\end{document}